# A Multilingual Parallel Corpora Collection Effort for Indian Languages

**Shashank Siripragada*†, Jerin Philip*†, Vinay P. Namboodiri‡, C.V. Jawahar†**
†International Institute of Information Technology - Hyderabad
‡Indian Institute of Technology - Kanpur
{shashank.siripragada@alumni,jerin.philip@research}.iiit.ac.in, vinaypn@iitk.ac.in, jawahar@iiit.ac.in

**Abstract**
We present sentence aligned parallel corpora across 10 Indian Languages - Hindi, Telugu, Tamil, Malayalam, Gujarati, Urdu, Bengali, Oriya, Marathi, Punjabi, and English - many of which are categorized as low resource. The corpora are compiled from online sources which have content shared across languages. The corpora presented significantly extends present resources that are either not large enough or are restricted to a specific domain (such as health). We also provide a separate test corpus compiled from an independent online source that can be independently used for validating the performance in 10 Indian languages. Alongside, we report on the methods of constructing such corpora using tools enabled by recent advances in machine translation and cross-lingual retrieval using deep neural network based methods.

## 1. Introduction and Related Work

Modern day neural network based approaches for Machine Translation (MT) are data hungry and sentence-level aligned parallel pairs are the currency. Neural MT (NMT) is currently the de-facto approach for training translation systems and shares the same traits (Koehn and Knowles, 2017).

Koehn (2005) uses European parliament proceedings available on the web to create evolving parallel corpora. The resource has been a major driving factor in attempts to build MT systems. We provide a similar effort through this work in providing multilingual parallel corpora for Indian languages.

Many languages spoken in the Indian subcontinent are categorized as low-resource (Post et al., 2012) considering the amount of parallel corpora available for training with deep neural network based models. There have been a few efforts towards addressing this lacuna. Among this, the IIT-Bombay Hindi English Parallel Corpus (IITB-hi-en) (Kunchukuttan et al., 2017) is the largest English-Hindi corpus available for training. Other significant efforts in constructing parallel corpora for Indian languages include Indian Language Corpora Initiative (ILCI) (Jha, 2010) across 10 Indian languages, restricted to health and tourism domains and Indic Multilingual Parallel Corpus (WAT-ILMPC) (Nakazawa et al., 2017), with possible shared content among multiple languages collected through automated efforts are limited and noisy (Philip et al., 2019). WAT-ILMPC is comprised of user contributed translations of subtitles, which are inadequate and often code-mixed. A specific effort is that of the Oriya-English Corpus (OdEnCorp) that provides a compilation in Oriya-English (Parida et al., 2020). Lack of resources reflects in the availability of automated translation systems of satisfactory quality in these languages. While some systems are reported (Wang et al., 2017; Philip et al., 2019) to perform well for English-Hindi, there are not many that generalize well for other languages. Based on these resources, in order to enable further progress in Indian languages, additional multi-lingual general purpose corpora that could be used for training over multiple Indian languages can be considered as a required resource. Further, in order to gauge the progress made in solving this task, it is also necessary to consider an independent general purpose test corpus that is not used for training. In this work, we address both these challenges.

Low resource languages have drawn an increased interest towards self-supervised extraction of sentence-aligned bitext (Schwenk, 2018; Schwenk et al., 2019; Artetxe and Schwenk, 2018) to augment training data for MT. Attempts at using pre-training with monolingual corpora and transferring the learning to machine translation have found moderate success (Guzmán et al., 2019). Another class of augmentation approaches (Edunov et al., 2018; Sennrich et al., 2016) turns to using noisy synthetic corpus obtained through back-translation to improve translation results in the low resource setting. However, the more relaxed the formulations are, the larger are the computational requirements for some improvement. Moreover, these formulations can be viewed as expectation maximization (EM) algorithms (Cotterell and Kreutzer, 2018; Graça et al., 2019) and often benefit from a good initialization - which in the case of MT is a collection of aligned sentences in two languages - a parallel corpus.

Another benefit of obtaining a parallel corpus is that it enables other resources such as cross-lingual word embeddings (Ruder et al., 2017) and sentence embeddings (Conneau et al., 2018) to also be developed. These resources would also enable downstream tasks such as paraphrase generation and question answering in multiple Indian languages that are currently less considered due to the lack of sufficient machine translation or word embedding capabilities.

All of this points to a strong need of usable parallel corpora for many languages to catch up. Recent literature suggests multilingual corpora with shared content among languages can provide significant boost to learning (Aharoni et al., 2019; Neubig and Hu, 2018; Johnson et al., 2017), enabling a single model to do zero shot translation and acting as regularization. Other neural approaches, which have further relaxed formulations include Schwenk et al. (2019). With many vernacular web sources publishing bilingual content and more speakers gaining access to the internet (Philip et al., 2019) - the resources available are growing. Combined with recent developments in natural language understanding with advent of deep neural networks - it is possible today to construct corpora with lesser human efforts.

---

*Equal contribution.

This work relies on two sources – the Press Information Bureau (PIB) and *Mann Ki Baat*, the Indian Prime Minister's speeches to compile sentence aligned parallel corpora with shared content across 10 languages. Our contributions are as follows:

- We revisit MT based alignment methods, to obtain the aligned parallel corpus with minimum human supervision.

- We release a corpus of size 407K sentence pairs compiled across 10 language pairs from PIB, intended as training data for multilingual models.

- We release an independent test corpus compiled from *Mann Ki Baat*. This is a useful resource to validate machine translation accuracy and generalization performance. The test-sets are between 9 languages with 2-3K example sentences on an average.

This document is structured as follows: §2. briefs the method involving NMT based alignment used to obtain sentence pairs. In §3., we summarize and describe the multilingual parallel corpora, the quality is validated by the evaluations elaborated in §4. We demonstrate state of the art performance in many tasks using the datasets in §4.

## 2. Aligned Parallel Corpora from Web

We take advantage of the structured information - links, dates etc. in the above sources of the same while crawling and pre-processing. Once we have alignments between documents, we use several strategies to obtain sentence level alignments across languages. On obtaining sentence level alignments, we use heuristics – length ratios, language identification through writing script etc, based filtering to remove noisy pairs.

**Crawling and Preprocessing** We crawl PIB and *Mann Ki Baat* websites for the articles. We extract only the text from html. Following the text extraction, we index and store the articles on the basis of a unique identification number that is assigned to each article. We additionally store metadata attributes of posted date and language the article was written in. This dataset of documents is subsequently processed to obtain delimited sentences in each document.

We use delimiter-based rules [1] specific to the language being processed to segment sentences in each language. For Urdu, we use UrduHack[2]. The sentences are tokenized for document-alignment and sentence-alignment by using SentencePiece (Kudo and Richardson, 2018) models trained by restricting each language to a vocabulary size of 4000 units.

**Document Alignments** For obtaining document level alignments in *Mann Ki Baat*, posted date of the article suffices as a retrieval criteria. The articles are posted in periodic manner with the multilingual content being uploaded with negligible time delay. Unlike *Mann Ki Baat*, PIB articles are posted with different timestamps making it harder to retrieve multilingual content.

Our method to align PIB documents closely resembles (Uszkoreit et al., 2010). Given an MT system, we can translate all non-English articles to English. This makes it possible to check similarity between two documents in English - one the original and the other a translation. We use cosine similarity on *term frequency-inverse term frequency* (TF-IDF), a measure commonly used to rank candidates in retrieval literature to obtain candidate articles. We restrict our search within the neighbourhood of 2 days of the posted date of source and choose the nearest neighbour to align documents. Using these procedures we obtain a document level alignment between articles in the various languages. From the pair of articles obtained through document alignment, we obtain sentence level alignments as described next.

**Sentence Alignments** We use BLEUAlign approach (Sennrich and Volk, 2010) where we have a reasonable working translation model between the language pairs. Sennrich and Volk (2010) proposed MT based sentence alignment algorithm denoted hereafter by BLEUAlign[3] which uses translation of either source or target text. It uses BLEU score (Papineni et al., 2002) as a similarity metric to obtain sentence level alignments. They demonstrate better alignments than conventional length based alignment methods (Gale and Church, 1993). The MT systems used in BLEUAlign are discussed next.

**NMT for Alignment** We use a single NMT model which is an implementation of the Transformer architecture (Vaswani et al., 2017) for a sequence-to-sequence learning task. This architecture consists of an encoder and a decoder. An encoder consumes source sequence represented by one-hot vectors corresponding to the tokens. The decoder uses the encoded representations of the source sequence, attends to it and autoregressively predicts the next token left to right. This constitutes the Transformer architecture. Parameters are shared among all languages for both the encoder and decoder (Johnson et al., 2017). The embeddings are also shared given the vocabulary at input and output are same through the multilingual formulation and for the benefits reported in Press and Wolf (2017). SentencePiece models help making training multiway models across 11 languages feasible without compromising coverage of all sentences in the language.

The dataset used to train our multilingual model is compiled from multiple sources: IIT Bombay English-Hindi corpus (Kunchukuttan et al., 2017), UFAL EnTam v2.0 (Ramasamy et al., 2012), OdiEnCorp (Parida et al., 2020), ILCI (Jha, 2010) and WAT-ILMPC (Nakazawa et al., 2017) multilingual corpora. Additionally we compile English-Telugu parallel text from Bible corpus (Christodouloupoulos and Steedman, 2015). We also augment training set with synthetic data obtained from Backtranslation (Sennrich et al., 2016) for Hindi and Telugu.

---

[1] https://github.com/jerinphilip/ilmulti
[2] https://github.com/urduhack

[3] https://github.com/rsennrich/Bleualign

|  |  | en | hi | ta | ur | ml | bn | te | or | gu | pa | mr |
|---|---|---|---|---|---|---|---|---|---|---|---|---|
| PIB (train) | Articles | 28K | 14.5K | 7.5K | 13.5K | 4K | 4.5K | 1K | 6K | 2.2K | 4.7K | 6.4K |
|  | Sentences | 1.3M | 440K | 227K | 360K | 97K | 112K | 22K | 89K | 60K | 88K | 165K |
|  | Aligned-en | - | 260K | 96K | 122K | 31K | 35K | 10K | 43K | 46K | 61K | 123K |
|  | Filtered | - | 156K | 61K | 45.3K | 17K | 21.6K | 6K | 9.1K | 25.5K | 26.3K | 40K |
|  | Vocabulary | 159K | 112K | 119K | 67.6K | 67K | 44.2K | 27K | 35K | 51.7K | 53.7K | 73K |
| Mann Ki Baat (test) | Articles | 58 | 57 | 47 | 8 | 45 | 48 | 48 | 39 | 45 | † | 47 |
|  | Sentences | 12.2K | 12.6K | 11.6K | 1.7K | 12.3K | 14.7K | 15.2K | 9K | 11.2K | - | 11.9K |
|  | Aligned-en | - | 5.4K | 6.0K | 1K | 5.2K | 5.8K | 5.5K | 0.8K | 6.8K | - | 6.2K |
|  | Filtered | - | 5.3K | 5.7K | 1K | 5K | 5.6K | 5.2K | 0.8K | 6.6K | - | 5.9K |
|  | Vocabulary | 21K | 11K | 23K | 4K | 22K | 17.5K | 21K | 4K | 19K | - | 20K |
|  | OOV rate | - | 13.3% | 26.4% | 14.1% | 36.9% | 39.7% | 74.1% | 28.3% | 30.4% | - | 16.8% |

Table 1: Detailed statistics of PIB and Mann Ki Baat. Number of sentences are reported after segmentation of articles. Aligned-en indicates number of sentences aligned to English across each language. Filtered indicates size after filtering aligned sentences through our pipeline. We also report vocabulary compiled from filtered sentences across languages. Details on nature of PIB and Mann Ki Baat corpus in sections 3.1. and 3.2. respectively. †Punjabi(pa) is unavailable in Mann Ki Baat.

| Source | #pairs | type |
|---|---|---|
| IITB-en-hi | 1.5M | en-hi |
| UFAL EnTam | 170K | en-ta |
| Backtranslated-Hindi | 2.5M | en-hi |
| WAT-ILMPC | 800K | xx-en |
| ILCI | 500K | xx-yy |
| OdiEnCorp | 27K | en-or |
| Telugu Bible | 30K | en-te |
| Backtranslated-Telugu | 500K | en-te |

Table 2: Training dataset used for multilingual model. xx-yy indicates parallel sentences aligned across multiple languages.

**Filtering aligned sentences for noise** Based on the two alignment techniques, we obtain a set of aligned candidate sentence pairs. Among the extracted sentence pairs through the methods described in the section above, we observe noisy sentences. These are in the form of URLs, numbers corresponding to dates, segments left untranslated in the news etc. We do not filter these early on as these can help in aligning sentences. However, to ensure that these do not affect training the machine translation model, we filter the data using `langid`[4] to remove foreign language tokens. We also filter based on the ratio of source-to-target length to filter imbalanced sentences amongst the pairs. BLEUAlign also has a filtering effect on sentence pairs as it can merge and discard sentences based on the BLEU score between target sentence and translation of source.

**Compiling Multilingual Content** Finally, to obtain aligned corpora across languages, we group sentence pairs extracted from many languages keyed by a single language. In this work, we use English as the common language. The choice is helped by the fact that many parallel corpora available for Indian languages have English as one of the two languages. Due to this all languages translate to English fluently. Through this procedure we obtain a multi-lingual parallel corpus.

## 3. Dataset Description and Statistics

In this section we describe the datasets produced with this work. Two datasets are compiled - the PIB corpus for training and *Mann Ki Baat* corpus for a standard test-set. Both corpora have shared sentences across multiple languages. We elaborate on the nature of the compiled corpora and the quality.

### 3.1. Press Information Bureau Corpus

Press Information Bureau (PIB) is an Indian government agency responsible for communications to media. These communications are released in the form of articles on a website[5] across 12 Indian Languages: Hindi (hi), Telugu (te), Tamil (ta), Malayalam (ml), Gujarati (gu), Kannada (kn), Urdu (ur), Bengali (bn), Oriya (or), Marathi (mr), Punjabi (pa), Assamese (as) and English (en). Out of these 12 Indian languages parallel corpora of variable sizes are available for 10 of them, excluding Assamese and Kannada. Articles are manually translated by PIB officials and can be deemed expert translations, which makes this a source for authentic parallel data. Articles are posted for about two years now since 2017. With progress of time more multilingual content will be added by the organization, which warrants improvements in size of corpus across all languages. These translations are adequate and fluent by nature, provided good alignment accuracy is obtainable. We observe that an article on an average provides 10 sentences.

**Corpus Statistics** We propose PIB corpora as a multilingual training set. The corpus compiled out of PIB contains

---
[4]https://github.com/saffsd/langid.py

[5]https://pib.gov.in

| Index | Corpus | Source | Target |
|---|---|---|---|
| 1 | PIB | इनमें से 3,18,931.22 करोड़ रुपये रक्षा बजट (रक्षा पेंशन के अतिरिक्त) के रूप में हैं। | Out of this Rs.3,18,931.22 crore has been earmarked for Defence (excluding Defence Pension). |
| 2 | PIB | इस अवसर पर विदेश मंत्रालय, सरकारी विभागों के वरिष्ठ अधिकारी, 38 पासपोर्ट अधिकारी और सेवा प्रदाता टीसीएस के अधिकारी उपस्थित थे। सर्वश्रेष्ठ कार्य करने वाले पासपोर्ट अधिकारियों और पुलिस अधिकारियों को बेहतर पासपोर्ट और सत्यापन के लिए उत्कृष्टता पुरस्कार दिए गए। | Senior officials of the ministry of External Affairs and other government departments beside 38 Passport officers and officials of service provider TCS attended the function where awards of excellence were presented to the best performing Passport Officers and Police Officers for best passport and verification services. |
| 3 | Mann Ki Baat | वर्षों पहले डॉ बाबा साहब आम्बेडकर ने भारत के औद्योगिकीकरण की बात कही थी। | Years ago, Dr. Baba Saheb Ambedkar spoke of Indias industrialization. |
| 4 | Mann Ki Baat | मैंने किसी को खादीधारी बनने के लिये नहीं कहा था । लेकिन मुझे खादी भण्डार वालों से जानकारी मिली कि एक सप्ताह में करीब करीब सवा सौ परसेन्ट हंड्रेड एंड ट्वेंटी फाइव परसेन्ट बिक्री में वृद्धि हो गयी । | I had not asked anyone to be Khadidhari, But the feedback I got from Khadi stores was that in a weeks time the sales had jumped up by 125. |

Table 3: Success cases of Alignment for sentence pairs from PIB and Mann Ki Baat. 1 and 3 are shorter sentences but relatively hard to align to English due to presence of sentence delimiters on the English side. 2 and 4 have longer Hindi sentence containing two segments delimited by *Purnavirama* (Hindi end of sentence marker) symbol. We observe that *BLEUAlign* merges these segments to obtain a higher BLEU score, aligning to a longer Hindi (1&3) and English (2&4) sentence which conveys same meaning as the source.

407K sentences aligned with shared content across multiple languages. More on the corpus statistics are described in Table 1. The articles in the corpus are aligned with English as a pivot language. Hindi and Urdu articles are being published since the start. Articles in the other languages are being brought up to speed recently leading to an imbalance between number of articles across languages (14.5K hi to 1K te). We also observe disproportion between number of articles and en-aligned sentences as not all articles posted have correspondence for English article in PIB. This leads to lesser number of aligned documents for extracting sentence pairs. The number of sentence pairs further decreases as BLEUAlign chooses to merge and discard source sentences based on BLEU score between translated source and target pairs. One such example is demonstrated in Table 3. Presence of large number of English tokens, numbers and duplicates among the aligned pairs results further reduction in size after filtering across languages. We observe variations in proportion of filtered sentences among languages. The variance in filtered sizes of the corpora is caused by erroneous matches by BLEUAlign which are pruned out by sentence-ratio based thresholding and example of which is present in Table 4. A lot many English words are also interspersed in Indian language articles.

**Multilingual shared content** Due to the presence of shared multilingual content among the languages in PIB we were able to extract aligned sentence pairs across multiple languages (xx-yy). The resulting corpora is described in Table 6 in format of a multilingual grid where we report sizes of sentence pairs aligned across languages. As PIB and *Mann Ki Baat* share all of the languages except one, we merge the obtained multilingual grid with upper triangle color coded Blue representing PIB corpora. en-hi have the highest number of parallel sentences constituting 38% of the corpora followed by en-ta and en-ur at 14% and 11% respectively. Because of these high proportions of en-aligned data we were able to obtain 22.3K hi-mr (related language pair), 26K hi-ta (distant language pair) sentence pairs by pivoting through english (refer Table 6). Such alignments among diverse set of languages can significantly boost the performance of multilingual NMT systems.

### 3.2. Mann Ki Baat

*Mann Ki Baat*[6] is a transcribed source for Indian Prime Minister's speeches across 12 Indian languages: Hindi, Telugu, Tamil, Malayalam, Gujarati, Kannada, Urdu, Bengali, Oriya, Marathi, Assamese, Manipuri (mp) and English. As the volume of this transcriptions is much less compared to PIB we compile a multilingual test. As these articles are relatively sparse[7], posted date serves as primary criterion for retrieving the multilingual content. These translations can be considered as expert translations and are fluent by nature as the speeches are addressed to people across India speaking many languages. These transcribed speeches usually tend to be longer, unlike PIB articles. *Mann Ki Baat* is also a promising source for multilingual corpora as more content is being added over time.

**Corpus Statistics** We release *Mann Ki Baat* as a multilingual test set upon which our models are evaluated. We extract sentences aligned with English as the pivot language compiling a bi-directional (xx-en) test set, described in Table 1. Volume of Urdu articles posted in *Mann Ki Baat* is relatively smaller resulting in fewer aligned sentences. Noise in the form of untranslated tokens and repetitions is also present in sentence aligned pairs of *Mann Ki Baat* although not as prevalent as in PIB resulting in removal of only a fraction of sentences after filtering.

---

[6] https://www.narendramodi.in/mann-ki-baat/
[7] Prime Minister's speeches happen in a periodic manner, usually with frequency of 1 per month.

| Index | Corpus | Source | Target |
|---|---|---|---|
| 1 | PIB | वो पीढ़ी दर पीढ़ी तैयार होते जाएंगे उसी क्षमता के साथ, सामर्थ्य के साथ तैयार होंगे। | They will be prepared about every generation. |
| 2 | PIB | रूपया 17 अप्रैल, 2017 को 64.41 रुपये प्रति अमेरिकी डॉलर के स्तर पर बंद हुआ। | 2017. Rupee closed at Rs. 64. |
| 3 | Mann Ki Baat | यहा के प्राथमिक विद्यालय के शिक्षक पी.के. मुरलीधरन और छोटी सी चाय की दुकान चलाने वाले पी.वी. चिन्नाथम्पी, इन दोनों ने, इस लाइब्रेरी के लिए अथक परिश्रम किया है | V. Chinnathampi who runs a small tea shop, have between them worked tirelessly for this library. |

Table 4: Failure cases of Alignment for sentence pairs from PIB and Mann Ki Baat. 1 represents a hard case to align, where the content is written elaborately in source and poorly in target language. 2 contains English sentence with punctuation, misaligned due to segmentation. 3 represents a case of partial alignment due to segmentation of the English sentence.

As *Mann Ki Baat* is released as a test set we compute OOV (Out-of-Vocabulary) rate for word types between PIB and *Mann Ki Baat* corpora (refer Table 1). We observe higher OOV rate among te (74.1%), bn (39.7%) as the size of English aligned corpus is less for these languages compared to others in PIB.

**Multilingual shared content** *Mann Ki Baat*, similar to PIB has shared multilingual content across languages. We extract sentence pairs (xx-yy). The resulting corpora is described in Table 6 in the format of a multilingual grid. Bottom triangle of the grid color coded in Red represents *Mann Ki Baat*. Presence of less number of en-aligned pairs among Urdu (ur) and Oriya (or) consequently led to less number of pairs across the grid for these two languages. For other languages we obtain test sets with sentences greater than 5K leading to significant number of aligned pairs (3K on an average) across all the languages in the grid. Test corpora of this size aligned across multiple language is previously unavailable, and can benefit standardization of multilingual models built for Indian languages.

### 3.3. Characterizations and Comparisons

Multilingual corpora available currently for Indian languages are ILCI (Jha, 2010) and WAT-ILMPC (Nakazawa et al., 2017). In this section we compare PIB and *Mann Ki Baat* with these multilingual corpus.

| Source | #sents | Vocab | #langs | Languages (xx) |
|---|---|---|---|---|
| ILCI | 550K | 640K | 11 | bn en hi ml ta te mr gu ur ka pa |
| WAT-ILMPC | 800K | 780K | 8 | bn en hi ml si ta te ur |
| PIB | 407K | 780K | 11 | bn en hi ml ta te mr gu ur pa or |
| *Mann Ki Baat* | 41K | 154K | 10 | bn en hi ml ta te mr gu ur or |

Table 5: Multilingual datasets available for Indian languages.

ILCI is a multilingual corpora with 50K sentence aligned pairs across 11 Indian languages and English and has marked a huge contribution in developing Translation systems for Indian languages since its release in 2010. However, ILCI is restricted to only health and tourism domain. In our past experiments, we observe that training on ILCI alone does not generalize well to real-world queries. ILCI dataset is only available for Indian nationals[8]. WAT-ILMPC is a multilingual corpora of subtitles collected from OPUS[9]. This corpus contains lot of short and untranslated segments. As the PIB corpus is translated by experts, we observe a rich vocabulary of words. In Table 5, we summarize the characteristics of these alongside PIB for comparison.

Test sets of multilingual nature available for Indian languages currently are aligned only to English (Nakazawa et al., 2017). We observe the similar problem of short and untranslated tokens in this test set which implies that good performance on this test set does not generalize good performance overall. *Mann Ki Baat* contains sentence alignments across 10 language pairs and English resulting in better understanding of translation phenomena across diverse set of Indian languages, some of them which are related (hi-mr, hi-gu) and some distant (hi-ta). *Mann Ki Baat* are also expert translations, containing a rich vocabulary. This test can be viewed as general domain as these are compiled from transcribed speeches addressing national crowd.

## 4. Evaluations

To demonstrate efficacy of the parallel corpus, we propose two methods of evaluation. First, we report the alignment quality across a small sample reported by humans. Second, we take two cases to check if the parallel corpora created are useful in Machine Translation - we train models one on the existing corpora alone and also use the newly obtained PIB corpus to augment existing corpora for training and demonstrate improvements in translation quality in the augmented case over the other.

### 4.1. Alignment Quality

We randomly sample 100 sentence pairs across all the languages of PIB and *Mann Ki Baat* after aligning to English. We manually evaluate the quality of alignments for these sentence pairs and report the number of sentences that have been verified as correct alignments in Table 7. In PIB, we observe high percentages for Urdu and Telugu indicating a good quality of alignment. *Mann Ki Baat* exhibits higher alignments accuracies due to a very strict one-to-one mapping between sentences in speeches and their English counterparts. On the other hand, PIB articles have relaxed writing, often merging or splitting sentences written in one language going to the other.

---
[8] http://www.statmt.org/wmt19/translation-task.html - Gujarati
[9] http://opus.nlpl.eu/

|    | en     | hi     | ta    | te   | ml    | ur    | bn    | gu    | mr    | or   | pa    |
|----|--------|--------|-------|------|-------|-------|-------|-------|-------|------|-------|
| en |        | 156344 | 60836 | 6035 | 17187 | 45355 | 21615 | 25598 | 40200 | 9101 | 26338 |
| hi | 5272   |        | 25848 | 1845 | 7865  | 9635  | 10476 | 13243 | 22378 | 2307 | 7751  |
| ta | 5744   | 2761   |       | 2275 | 5327  | 3416  | 6704  | 8575  | 12180 | 974  | 4778  |
| te | 5177   | 2289   | 3100  |      | 1196  | 564   | 917   | 1909  | 829   | 183  | 1086  |
| ml | 5017   | 2305   | 3124  | 2898 |       | 1220  | 3243  | 4309  | 3628  | 376  | 1921  |
| ur | 1019   | 742    | 637   | 599  | 624   |       | 1631  | 2100  | 1500  | 741  | 5214  |
| bn | 5634   | 2706   | 3460  | 2939 | 2938  | 559   |       | 4718  | 4618  | 490  | 1906  |
| gu | 6615   | 3213   | 3998  | 3528 | 3469  | 749   | 3810  |       | 5580  | 620  | 3702  |
| mr | 5867   | 2491   | 3175  | 2839 | 2803  | 490   | 3054  | 3658  |       | 867  | 3080  |
| or | 768    | 389    | 470   | 440  | 427   | 98    | 447   | 541   | 432   |      | 889   |
| pa[†] | -   | -      | -     | -    | -     | -     | -     | -     | -     | -    |       |

Table 6: Multilingual shared content across language pairs for PIB (Blue) and Mann Ki Baat (Red). Rows and columns indicate language pairs. Upper triangle (Blue) indicates PIB across Languages. Lower triangle (Red) indicates Mann Ki Baat. The intensity of the cell color is proportional to size of the sentences aligned pairs. † Mann Ki Baat does not contain Punjabi.

In Oriya, our translation models do not perform as well, evident from NMT performance reported in Table 8 which consequently leads to poor retrieval based alignments. We could not observe many articles we could align with pure date based matches either. Mismatch in document alignments further cause noisy sentence alignments for the case of PIB. With the progress of time more data will be made available on the PIB website leading to better mapping of Oriya articles and subsequently better alignments. This will improve the performance of NMT models. In current state, we provide the corpus except for Oriya where the alignment accuracy is observed to be reasonably good. However, for *Mann Ki Baat*, we observe that despite the same model, sentences are aligned well. The articles matching correctly and the existence of a strict one to one mapping in the sentences significantly helps in this case.

| Language Pairs | PIB | Mann Ki Baat |
|---|---|---|
| en-hi | 94% | 99% |
| en-ta | 94% | 98% |
| en-te | 97% | 100% |
| en-ml | 93% | 100% |
| en-ur | 96% | 100% |
| en-bn | 87% | 99% |
| en-mr | 87% | 99% |
| en-or | 2% | 95% |
| en-gu | 91% | 100% |
| en-pa | 90% | N/A |

Table 7: Assessment of sentence alignment quality.

Qualitative examples for English-Hindi demonstrating success and failure of alignment pipeline are presented in Tables 3 and 4. We are able to align and extract even significantly long and complex sentences from PIB and Mann Ki Baat. Among the failure cases in Table 4, where a fragment matches to a larger sentence due to an alignment error, the possibility to bring down errors of this sort by heuristics like length ratio can be observed.

### 4.2. Translation Quality

In order to evaluate translation quality, we train a standard NMT system (Philip et al., 2019) across all language pairs in a multilingual NMT. These are trained for various cases such as benchmarking training with ILCI or with PIB. We use Bilingual Evaluation Under Study (BLEU) (Papineni et al., 2002) a modified precision based score used in sequence-to-sequence tasks to compare the models trained by different variations of data. We report the BLEU scores obtained using Indic NLP Library[10]. We first consider the multilingual model across languages that we use. We next consider the utility of the proposed PIB corpora. In all our evaluations we also consider evaluating the model with the proposed new *Mann Ki Baat* test set.

**Multilingual scores** Multilingual models enable zero-shot translation (Johnson et al., 2017). They can translate among unseen pairs of languages during inference which are unseen at training time by implicitly pivoting through shared content. This enables reporting BLEU scores across languages in multilingual grid format in Table 8, given the presence of a test set - *Mann Ki Baat* with samples in all directions. Similarly, we report the scores on ILCI with a random subset held out from the training set. Training of all the languages is done on the specified corpora and test is done on the ILCI and *Mann Ki Baat* test set.

### 4.3. PIB for NMT

In order to check the usefulness of PIB corpora, we compute the BLEU scores of the model trained with all existing data and the model trained on the same data augmented with PIB corpora. We further consider the cross-domain generalization ability of models by training individual multilingual models and assess it on existing WAT-ILMPC, ILCI and the proposed new *Mann Ki Baat* corpora. Due to unavailability of separate evaluation set in ILCI we randomly sample a test set containing 500 sentences.

---
[10] https://anoopkunchukuttan.github.io/indic_nlp_library/

|    | ILCI | | | | | | | | | | Mann Ki Baat | | | | | | | | | |
|----|------|------|------|------|------|------|------|------|------|------|------|------|------|------|------|------|------|------|------|------|
|    | en | hi | ta | te | ml | ur | bn | gu | mr | pa | en | hi | ta | te | ml | ur | bn | gu | mr | or |
| en | 0.0 | 30.0 | 5.17 | 9.18 | 5.59 | 23.5 | 14.3 | 21.8 | 13.6 | 22.7 | 0.0 | 13.8 | 2.44 | 2.94 | 2.51 | 8.08 | 4.63 | 7.23 | 4.8 | 0.76 |
| hi | 29.0 | 0.0 | 8.12 | 16.1 | 8.78 | 58.6 | 21.3 | 48.3 | 26.7 | 54.0 | 17.8 | 0.0 | 2.81 | 4.82 | 4.02 | 41.2 | 7.62 | 24.3 | 10.4 | 0.7 |
| ta | 13.6 | 19.2 | 0.0 | 7.43 | 5.36 | 16.5 | 9.72 | 15.0 | 9.23 | 16.3 | 9.7 | 11.1 | 0.0 | 3.41 | 2.95 | 10.1 | 3.23 | 5.61 | 3.68 | 0.49 |
| te | 19.8 | 28.7 | 5.62 | 0.0 | 6.61 | 24.0 | 13.6 | 22.8 | 13.8 | 22.0 | 7.19 | 11.4 | 2.18 | 0.0 | 3.22 | 10.6 | 3.58 | 6.13 | 4.14 | 0.3 |
| ml | 17.0 | 23.3 | 5.78 | 10.0 | 0.0 | 19.4 | 13.0 | 18.9 | 11.3 | 18.9 | 11.2 | 14.5 | 2.43 | 4.25 | 0.0 | 12.3 | 4.24 | 7.36 | 4.72 | 0.31 |
| ur | 26.0 | 58.9 | 6.65 | 13.6 | 8.04 | 0.0 | 17.3 | 39.0 | 20.2 | 41.5 | 15.1 | 39.6 | 1.95 | 3.89 | 3.27 | 0.0 | 6.06 | 18.4 | 7.7 | 0.41 |
| bn | 20.8 | 31.3 | 5.83 | 11.5 | 7.21 | 24.5 | 0.0 | 24.2 | 14.7 | 22.8 | 10.8 | 16.2 | 1.82 | 3.51 | 2.76 | 14.1 | 0.0 | 8.92 | 5.13 | 0.41 |
| gu | 27.8 | 57.3 | 7.35 | 15.1 | 8.67 | 45.4 | 20.3 | 0.0 | 23.8 | 40.9 | 14.6 | 37.3 | 2.26 | 4.55 | 3.61 | 29.5 | 7.02 | 0.0 | 9.73 | 0.5 |
| mr | 24.0 | 43.8 | 6.7 | 13.2 | 8.09 | 33.9 | 17.8 | 33.0 | 0.0 | 32.1 | 11.5 | 19.5 | 2.11 | 3.59 | 3.33 | 17.9 | 5.08 | 12.7 | 0.0 | 0.53 |
| xx | 29.2 | 70.8 | 7.51 | 15.0 | 9.13 | 55.3 | 20.0 | 45.4 | 24.5 | 0.0 | 3.12 | 8.41 | 1.38 | 1.46 | 1.15 | 8.89 | 3.19 | 4.89 | 3.11 | 0.0 |

Table 8: BLEU scores of model trained on unaugmented data on multilingual test sets - test-split sampled from ILCI and Mann-Ki Baat obtained from the unaugmented model. Rows correspond to source languages and columns target languages. xx is (Punjabi and Oriya respectively) depending on ILCI or Mann Ki Baat.

As can be observed from Table 8, training on unaugmented corpora alone results in a reasonable performance on ILCI test but a deficient performance on the new *Mann Ki Baat* test set. This indicates that the *Mann Ki Baat* test set is more challenging as compared to the ILCI test set. Further, the performance of this model on these test sets is not as good as they are being evaluated in a cross-domain setting.

When we consider the performance of PIB as an additional corpora used along with unaugmented model, the results are significantly improved. The results for the same are provided in Table 10. As can be observed, the results in this case show significant improvement in BLEU scores in most cases. The only language in which we see a small decrease in this setting is in Oriya for *Mann Ki Baat* corpora. This we believe is due to the relatively few sentences obtained for Oriya and the fact that the baseline NMT system for Oriya language is noisy.

Table 8 additionally provides insights into the effectiveness of multilingual model trained on comparable languages. Translation across strongly related languages are higher, as the mapping is easier to learn. An example we observe is Hindi, Gujarati and Marathi that belong to Indo-Aryan family[11]. Pairs among these languages have high BLEU scores. We also observe high BLEU scores among Hindi, Urdu as it is generally regarded that Urdu and Hindi are quite similar phonetically only differing in writing script[12]. This shared linguistic aspects are very helpful as improving translation numbers in one language in the family helps improving across the family due to multilingual models.

We finally consider cross-domain generalization performance to demonstrate the efficacy of PIB corpus. In Table 9 we report cross domain inference scores of multilingual NMT models trained on ILCI (1), WAT-ILMPC (WAT)(2) and PIB (3) and tested on the remaining corpora respectively. We additionally consider Mann-Ki-Baat (MKB) test corpus. On ILCI test domain, model 3 achieves +8.7 BLEU in en-hi direction when compared to 2. Similar increments can be found in the case of WAT and MKB test domains with 3 outperforming 1 and (1,2) respectively, indicating model trained on PIB (3) transfers well to other domains. The diagonal values in case of 1 and 2 are high as in-domain performance of the model is expected to be higher.

---

[11] https://en.wikipedia.org/wiki/Indo-Aryan_languages
[12] https://www.britannica.com/topic/Urdu-language

| id | train \ test | ILCI | WAT | MKB |
|----|------|------|-----|-----|
| 1 | ILCI | 21.7 | 6.15 | 6.52 |
| 2 | WAT | 5.98 | 25.4 | 5.1 |
| 3 | PIB | 14.7 | 7.15 | 15.2 |

Table 9: English-Hindi BLEU scores for cross domain inference. Rows indicate training corpus and columns indicate test corpus across different domains.

## 5. Conclusion and Future Work

In this paper we have presented parallel corpora for 10 Indian languages and demonstrated its effectiveness in multilingual machine translation. Our systems are set to collect more sentence pairs as the more articles are published, similar to Europarl (Koehn, 2005). The nature of the corpora opens avenues to explore on how to efficiently use weakly aligned formulations to improve the corpora iteratively.

Many languages that are related have better translation performance between them compared to English, examples being Hindi-Punjabi, Hindi-Marathi, Hindi-Urdu. MT based alignments between these can be used to create larger mappings amongst the sentences in both. Iterative application of such methods can lead to a very refined corpus, through minimum human annotation.

Finally, we also provide a general test corpora that can be used to independently evaluate MT performance for Indian Languages in a generic setting.

## 6. Release Information

Through this work release a multilingual parallel corpora that is obtained from online publicly available documents. The release comprises 407K sentences across 11 languages

Mann Ki Baat

|    | en   | hi   | ta   | te    | ml   | ur    | bn   | gu   | mr   | or    |
|----|------|------|------|-------|------|-------|------|------|------|-------|
| en | 0.00 | 2.10 | 1.60 | 0.85  | 2.10 | 9.42  | 2.76 | 3.77 | 3.20 | -0.53 |
| hi | 2.40 | 0.00 | 2.74 | 0.61  | 2.49 | -3.12 | 3.48 | 5.80 | 4.40 | -0.42 |
| ta | 3.90 | 3.40 | 0.00 | -0.42 | 1.56 | 2.76  | 3.08 | 3.24 | 2.92 | -0.37 |
| te | 3.01 | 2.80 | 1.57 | 0.00  | 1.48 | 1.85  | 2.56 | 2.71 | 2.72 | -0.30 |
| ml | 3.90 | 2.40 | 2.06 | -0.06 | 0.00 | 3.69  | 3.00 | 3.14 | 3.19 | -0.31 |
| ur | 6.00 | 2.80 | 1.81 | 0.45  | 2.27 | 0.00  | 3.92 | 4.70 | 4.30 | -0.41 |
| bn | 4.10 | 2.40 | 1.88 | 0.05  | 2.06 | 3.60  | 0.00 | 3.58 | 3.14 | -0.33 |
| gu | 5.20 | 2.80 | 2.54 | 0.17  | 2.47 | 3.78  | 3.88 | 0.00 | 4.07 | -0.50 |
| mr | 4.60 | 3.30 | 2.23 | 0.42  | 1.76 | 2.76  | 3.94 | 3.50 | 0.00 | -0.33 |
| or | 8.48 | 6.09 | 2.07 | 0.13  | 2.33 | -0.66 | 3.95 | 4.68 | 2.47 | 0.00  |

|    | en    | hi   | ta    | te    | ml    | ur    | bn    | gu   | mr   | pa   |
|----|-------|------|-------|-------|-------|-------|-------|------|------|------|
| en | 0.00  | 1.40 | -0.42 | 0.39  | -0.12 | 1.30  | 0.60  | 1.30 | 1.20 | 2.20 |
| hi | 0.70  | 0.00 | 0.53  | -0.10 | 0.37  | 0.40  | 0.30  | 0.20 | 1.30 | 1.80 |
| ta | 0.20  | 0.40 | 0.00  | 0.52  | 0.34  | -0.20 | 0.57  | 1.40 | 0.96 | 1.50 |
| te | 1.30  | 1.70 | -0.03 | 0.00  | 0.55  | 0.30  | 1.40  | 2.40 | 1.70 | 1.50 |
| ml | 0.10  | 1.00 | -0.17 | 0.02  | 0.00  | -0.30 | -0.10 | 0.70 | 0.50 | 0.60 |
| ur | 0.70  | 0.40 | 1.33  | 0.70  | 0.06  | 0.00  | 0.20  | 1.40 | 0.80 | 1.50 |
| bn | 0.90  | 0.50 | 0.22  | -0.20 | -0.16 | 0.80  | 0.00  | 0.70 | 1.00 | 1.80 |
| gu | 1.00  | 0.00 | 0.20  | 0.20  | -0.12 | -0.40 | 0.60  | 0.00 | 0.40 | 1.50 |
| mr | 1.10  | 1.10 | 0.53  | 0.10  | -0.25 | 0.30  | 0.40  | 1.30 | 0.00 | 1.70 |
| pa | -0.10 | 0.30 | 0.04  | 0.30  | -0.02 | 0.20  | 0.50  | 1.00 | 0.50 | 0.00 |

ILCI

Table 10: Improvements (differences in BLEU) in directions by augmenting with PIB vs un-augmented with all remaining data. Reds indicate drop in BLEU scores and blues indicate improvements. We see improvements overall by using PIB as augmented data for training.

for the PIB corpus for training and 41K sentences across 10 languages for *Mann Ki Baat* corpus for test. This is the first release of the corpus. The corpora is available for download at http://preon.iiit.ac.in/~jerin/bhasha. As these online resources would be increasing, particularly in low resource Indian languages such as Kannada and Punjabi, this work in corpus construction can be further extended for future releases.

## Acknowledgements

We gratefully acknowledge the online corpora provided by ILCI, WAT-ILMPC, the online publicly available data that we have used in this work and the online open source tools that have facilitated this work.